\documentclass{article}

\usepackage{PRIMEarxiv}

\usepackage[utf8]{inputenc} 
\usepackage[T1]{fontenc}    
\usepackage{hyperref}       
\usepackage{url}            
\usepackage{booktabs}       
\usepackage{amsfonts}       
\usepackage{nicefrac}       
\usepackage{microtype}      
\usepackage{lipsum}
\usepackage{fancyhdr}       
\usepackage{graphicx}       
\graphicspath{{media/}}     

\usepackage{amsmath}
\usepackage{amssymb}

\usepackage{nicefrac}       
\usepackage{xcolor}         
\usepackage{bm}
\usepackage{multirow}
\usepackage{diagbox}
\usepackage{colortbl}
\usepackage{pifont}
\usepackage{enumitem}

\title{Exploit CAM by itself: Complementary Learning System for Weakly Supervised Semantic Segmentation}

\author{
  Jiren Mai\\
  Southeast University \\
  Nanjing, China\\
  \texttt{maijiren@seu.edu.cn} \\
   \And
  Fei Zhang \\
  Shanghai Jiao Tong University \\
  Shanghai, China\\
  \texttt{ferenas@sjtu.edu.cn} \\
   \And
  Junjie Ye, Marcus Kalander\\
  Huawei Noah’s Ark Lab\\
  Shenzhen \& Hong Kong, China\\
  \texttt{\{yejunjie4, marcus.kalander\}@huawei.com} \\
   \And
  Xian Zhang, Wankou Yang \\
  Southeast University \\
  Nanjing, China\\
  \texttt{\{zzx\_ovo, wkyang\}@seu.edu.cn} \\
   \And
  Tongliang Liu \\
  The University of Sydney \\
  Sydney, Australia\\
  \texttt{tongliang.liu@sydney.edu.au}\\
   \And
  Bo Han \\
  Hong Kong Baptist University \\
  Hong Kong, China\\
  \texttt{bhanml@comp.hkbu.edu.hk}\\
}

\begin{document}
\maketitle

\begin{abstract}
Weakly Supervised Semantic Segmentation (WSSS) with image-level labels has long been suffering from fragmentary object regions led by Class Activation Map (CAM), which is incapable of generating fine-grained masks for semantic segmentation. To guide CAM to find more non-discriminating object patterns, this paper turns to an interesting working mechanism in agent learning named Complementary Learning System (CLS). CLS holds that the neocortex builds a sensation of general knowledge, while the hippocampus specially learns specific details, completing the learned patterns. Motivated by this simple but effective learning pattern, we propose a General-Specific Learning Mechanism (GSLM) to explicitly drive a coarse-grained CAM to a fine-grained pseudo mask. Specifically, GSLM develops a General Learning Module (GLM) and a Specific Learning Module (SLM). The GLM is trained with image-level supervision to extract coarse and general localization representations from CAM. Based on the general knowledge in the GLM, the SLM progressively exploits the specific spatial knowledge from the localization representations, expanding the CAM in an explicit way. To this end, we propose the Seed Reactivation to help SLM reactivate non-discriminating regions by setting a boundary for activation values, which successively identifies more regions of CAM. Without extra refinement processes, our method is able to achieve breakthrough improvements for CAM of over 20.0\% mIoU on PASCAL VOC 2012 and 10.0\% mIoU on MS COCO 2014 datasets, representing a new state-of-the-art among existing WSSS methods. 
\end{abstract}


\section{Introduction}

Semantic segmentation plays an important role in computer vision, which aims to classify each pixel in an image. Due to the success of deep learning and CNNs, semantic segmentation has witnessed great progress in recent years, giving birth to numerous remarkable works \cite{DeeplabV2, DeeplabV3, Huang_2019_ICCV, strudel2021segmenter, cheng2021per}. However, these methods heavily rely on accurate pixel-wise annotations for fully supervised training, which is time-consuming and labor-intensive to collect, making them less economical to apply. Weakly-supervised semantic segmentation (WSSS) is developed to liberate humans from these exhaustive annotation efforts, using weaker and cheaper annotations to achieve semantic segmentation. Image-level labels~\cite{IRN, CPN, L2G, W-OoD}, scribbles~\cite{Sribble}, bounding boxes~\cite{Box1, Box2}, and points~\cite{Point} are some annotation types commonly used for WSSS. Especially, this paper focuses on WSSS based solely on image-level labels.

A key problem in WSSS is how to derive localization cues with only the supervision of image-level labels. By exploiting the contribution of local regions to classification confidences, Class Activation Maps~\cite{CAM} (CAM) provides a key idea and has been commonly used in current WSSS methods~\cite{SEC,AMR,AdvCAM,CPN,CONTA,SEAM}. However, CAM is incapable of serving as a fine-grained mask, as it simply captures the small salient regions with discriminating features. To address this issue, many works~\cite{SEAM, DRS, CONTA, CPN} have tried to drive the CAM to cover more of the target regions. These methods have achieved some success, but the artificially designed modules are complex and it is difficult to integrate their advantages. In this paper, we turn to an interesting concept in agent learning theories, improving CAM based on an experiential learning process.

\begin{figure*}
  \centering
  \includegraphics[width=0.9\textwidth]{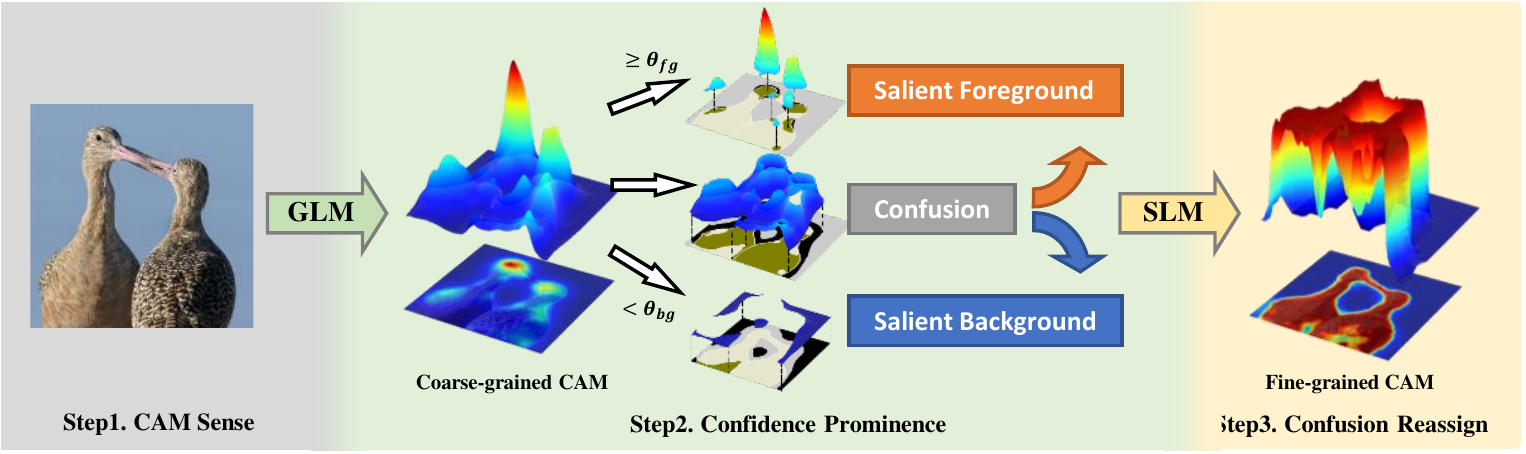}
  \caption{Overview of our method. The General Learning Module (GLM) senses coarse-grained CAM, which is divided into three parts by thresholds of $\theta_{bg}, \theta_{fg}$, and the corresponding local ground-truth is shown below (brown for foreground and black for background). With the guidance of the marked CAM, the Specific Learning Module (SLM) supports confusion reassigning to build a fine-grained CAM.}
  \label{overview}
\end{figure*}

The Complementary Learning Systems (CLS)~\cite{CLS1995, CLS2016} hypothesis suggests that two learning systems play complementary roles between the neocortex and the hippocampus in human learning process. The neocortex misses some details but generally builds a perception of patterns in target knowledge, while the hippocampus repeatedly consolidates learned patterns and completes details. We contend that the problem of the non-discriminating regions overlooked by CAM is similar to that of the missing details in the neocortex. As shown by the confidence distribution of the CAM in Figure~\ref{overview}, most pixels in the CAM with neutral confidence are difficult to assign as foreground or background. Indeed, the confusion regions represented by these pixels are mixed with true foreground (brown pixels in the figure) and true background (black pixels). As the salient foreground and background are what CAM has learned, we treat the confusion regions as the missing details of CAM. Therefore, we propose a General-Specific Learning Mechanism (GSLM) to help CAM reactivate the missing details and reassign these confusion regions, similar to the collaborative work of the neocortex and hippocampus in CLS. In GSLM, a General Learning Module (GLM) learns patterns of image-level labels and provides localization representations. A Specific Learning Module (SLM) inherits GLM and adjusts the network weights with these representations, supported by the proposed activation loss Eq.~\ref{act_loss}. The representations are termed Confidence CAM and are calculated from the original CAM by our proposed Coarse Generation Eq.~\ref{coarse-generation}, which marks salient and confusion regions with boundary constraints. To achieve the reassignment of the confusion regions, Seed Reactivation Eq.~\ref{wam} is integrated into SLM, which uses bounded functions to suppress salient regions and activates non-discriminating regions. By applying GSLM, a fine-grained CAM can be directly obtained by the conventional CAM network. In addition, we show that GSLM is compatible with other CAM generators~\cite{SEAM, AdvCAM, CPN, IRN}, and can thus be applied to advanced methods for further performance breakthroughs.

We set up performance experiments on the PASCAL VOC 2012~\cite{voc} and the MS COCO 2014~\cite{coco} datasets to verify the effectiveness of our method. For accuracy of CAM, our method improves the baseline (48.6\% mIoU) by 22.1\% mIoU on the PASCAL VOC 2012 $train$ set and the baseline (32.5\% mIoU) by 10.8\$ mIoU on the MS COCO 2014 $train$ set, ahead of existing methods over 11.6\% mIoU. For the accuracy of pseudo-masks, our method achieves 75.1\% mIoU on the PASCAL VOC 2012 $train$ set and 43.7\% mIoU on the MS COCO 2014 $train$ set, which is far beyond existing methods. For segmentation results, our method achieves a new state-of-the-art performance of 70.6\% mIoU both on $val$ and $test$ set of PASCAL VOC 2012 and 40.9\% mIoU on the MS COCO 2014 $val$ set.

Our main contributions are summarized as follows:
\begin{itemize}[leftmargin=*]

\item We propose a General-Specific Learning Mechanism (GSLM), a simple yet efficient training process to drive CAM to reassign the confusion regions, producing fine-grained CAM.

\item We apply GSLM to different advanced methods and achieve further improvements in the accuracy of the CAMs.

\item Experimental results on both PASCAL VOC 2012 and MS COCO 2014 show that our method outperforms the previous state-of-the-art. In particular, our method boosts baseline CAM by 22.1\% mIoU the PASCAL VOC 2012 $train$ set, ahead of the best existing methods over 11.6\% mIoU.
\end{itemize}

\section{Related Work}
\label{related}

\paragraph{Semantic Segmentation.}

Semantic segmentation aims at assigning a predefined category to every pixel on a given image. After introducing the fully convolutional network (FCN)~\cite{Long_2015_CVPR} into this task, researchers have designed various efficient models to improve performance, including dilated convolutions\cite{DeeplabV2}, encoder-decoder architecture\cite{DeeplabV3, ronneberger2015u}, and feature pyramid\cite{zhao2017pyramid, DeeplabV3}. To further promote the model's ability of context aggregation\cite{zhang2019co, yuan2021ocnet}, self-attention\cite{Huang_2019_ICCV, Li_2019_ICCV} paradigms are utilized to learn long-range dependency. Based on this, vision transformers~\cite{strudel2021segmenter, cheng2021per} are adapted to segmentation tasks, which occupy the state-of-the-art model in different benchmarks. This paper focuses mainly on semantic segmentation in the weakly supervised scenario.

\paragraph{Weakly Supervised Semantic Segmentation.}

Weakly Supervised Semantic Segmentation (WSSS) aims to generate pixel-level annotation for segmentation tasks. For Image-level based WSSS, most works~\cite{SEC,AMR,AdvCAM,CPN,CONTA,SEAM} have followed a prevailing pipeline to address WSSS, which could be described as 1) training a pseudo-mask generator with image-level labels, and 2) training a fully-supervised semantic segmentation with the generated pseudo-mask labels. Clearly, generating an initial seed map to serve as the pseudo-mask (step 1) is vital for WSSS. To achieve this goal, nearly all works have turned to CAM, which mark out the localization information of an image by merely using a classification network. However, CAM merely focuses on incomplete and fragmentary object regions, driving a gap toward the fine-grained target segmentation maps. To address this issue, there are two mainstream methods to help CAM extract more potential seeds. 

The first category mainly focuses on generating better seed regions during the generation of CAM~\cite{SEAM, CONTA, CPN, AdvCAM, SIPE, DRS, OAA}. Some methods~\cite{AE-PSL, CGNet} focused on erasing the discriminating areas or features in the classification network~\cite{SeeNet, FickleNet}, forcing CAM to pay attention to discriminate other potential object areas. \cite{AE-PSL} proposed to deliberately erase specific regions recognized from CAM, and iteratively re-train them for completing object region. Another idea is to artificially add prior knowledge to lead the CAM network to pay attention to potential features. Cross-image mining methods~\cite{CIAN, Co-attention} added cross-image modules that collect semantics on a higher level, from the relation of multiple images with the same classes. Besides, global memory methods \cite{FAM, OAA} turned to constantly record the category features during the training period, expanding the CAM based on the knowledge learned in previous stages. Furthermore, some works~\cite{SEAM,CPN} investigated specific data regularizations to lead the CAM expansion in an explicit way. SEAM~\cite{SEAM} proposed to leverage the scale invariance of the spatial characteristics while CPN~\cite{CPN} turned to help CAM by using a complementary pair of image inputs.

The second category investigates post-processing to refine the well-trained CAM. Conditional Random Fields (CRF)~\cite{CRF} is one of the most frequently used non-training refiner methods in WSSS. To take full advantage of the pixel-level semantic relationship, PSA~\cite{PSA} and IRN~\cite{IRN} aimed to learn the similarity between pixels obtained from CRF, and apply random walk to further refine the seed areas.

\section{Preliminaries \& Motivation}
\label{motivation}

\paragraph{Class Activation Map.}

In the prevailing pipelines~\cite{SEC, CONTA}, WSSS methods firstly generate localization seeds from image-level labels, which are then refined into pseudo-masks for fully supervised semantic segmentation network. Intuitively, the initial seed maps in the first stage is vital to WSSS. To generate the seed maps, most methods resort to CAM, which is an efficient method that extracts localization maps from a classification network. Specifically, the classification network for CAM consists of a feature extract backbone, a Global Average Pooling (GAP) and a classification layer. Given the input image $x$ of size $3 \times H \times W$, with a height $H$, width $W$, and $C$ potential categories, we denote the combination of the feature extract backbone and the classification layer as $f: \mathbb{R}^{3 \times H \times W} \rightarrow \mathbb{R}^{C \times H \times W}$. The classification prediction score $\hat{y} \in \mathbb{R}^C$ can be calculated as $\hat{y} = \sigma(\text{GAP}(f(x)))$, where $\sigma$ denotes the sigmoid activation function. To train the network, the classification loss is defined as the binary cross-entropy between prediction $y$ and ground-truth $\hat{y}$, i.e.,

\begin{gather}
\label{cls_loss}
L_\text{cls}(\hat{y}, y) = -\frac{1}{C}\sum^C_c{(y_c\ln{\hat{y}_c}+(1-y_c)\ln{(1-\hat{y}_c)})}
\end{gather} 

After training $f$, the CAM of an image $x$ related to the $c$-th class, denoted as $M_c \in \mathbb{R}^{C\times H \times W}, c \in \{1,2,..,C\}$, can be directly calculated as,
\begin{gather}
\label{cam}
M_c = \frac{ReLU(f(x)_c))}{\max f(x)_c}.
\end{gather}

\paragraph{Shortage of CAM.}

\begin{figure}
  \centering
  \includegraphics[width=0.43\textwidth]{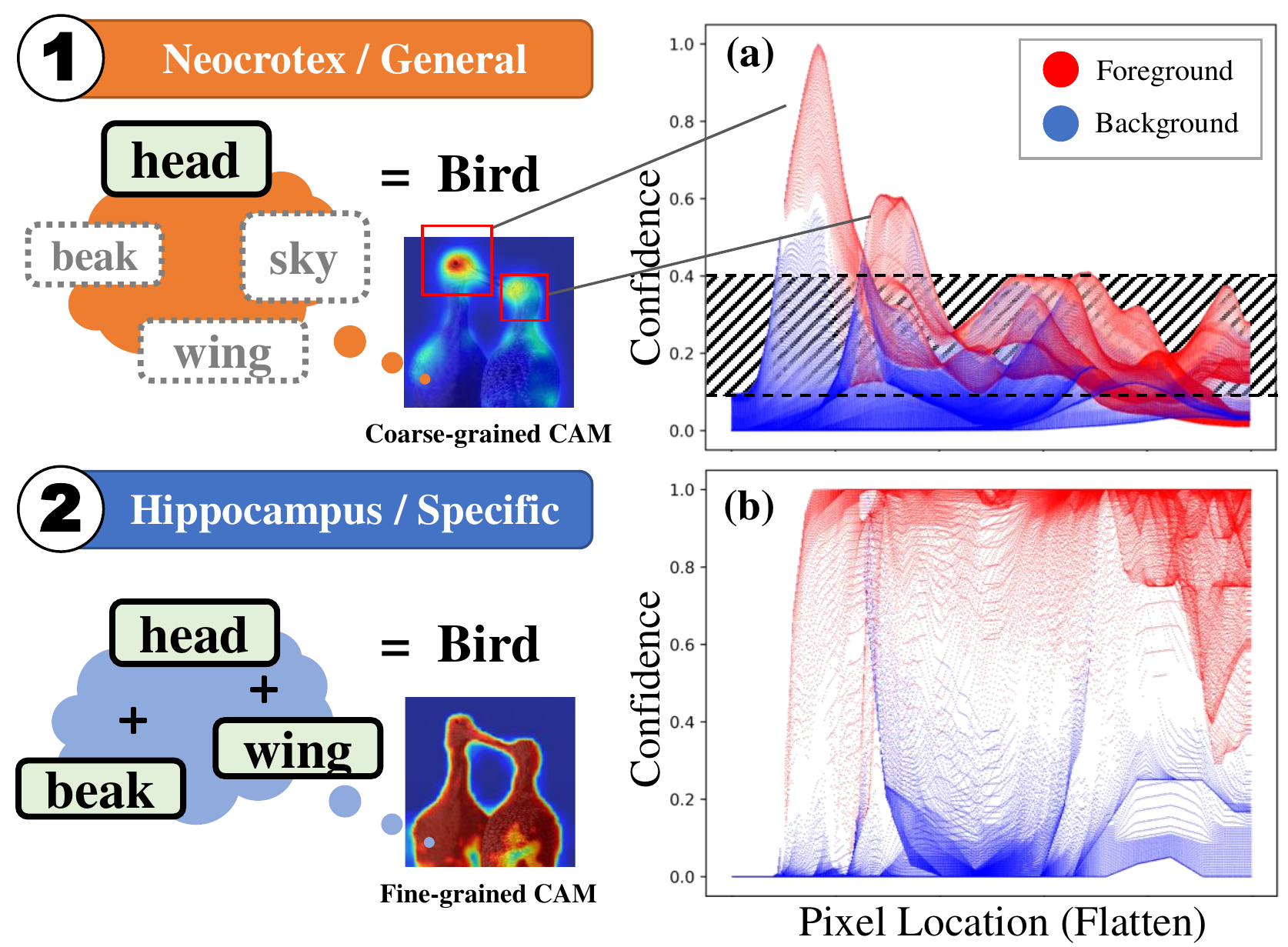}
  \caption{Learning stages of CLS and GSLM. For target patterns of birds, \ding{172} neocortex / General Learning Module (GLM) builds a general sense of the target, and \ding{173} hippocampus / Specific Learning Module (SLM) further learns details for completing patterns. The distribution of confidence of (a) Coarse-grained CAM and (b) Fine-grained CAM produced from two stages are shown in right.}
  \label{motivation-figure}
\end{figure}

In principle, CAM roughly indicates the contribution of local image regions to classification confidence~\cite{IRN}. The discriminating features always play a dominant role in the classification, as can be seen from Figure~\ref{motivation-figure}(a), where the confidence of the bird head regions (highlighted by red boxes) is much ahead of other regions. The phenomenon results in CAM mainly confined to the small salient regions. Most methods~\cite{SEAM, DRS, CONTA, CPN} modify the structure of the CAM generating network or introduce well-designed modules, which alter the contribution of image regions to classification confidence, leading to the expansion of the CAM. However, Figure~\ref{motivation-figure}(a) suggests that foreground points (in red) generally lie above background points (in blue), indicating that CAM still treats non-discriminating features differently from the background. Here, we pose the question \textit{Is CAM fully exploited?}

\paragraph{Fully exploit CAM.}

We hold that CAM is not fully exploited by simply training a classification network. As shown by the shaded domain in Figure~\ref{motivation-figure}(a), most pixels with neutral confidence in CAM are difficult to accurately assign to either foreground or background, representing ``confusion regions''. For a classification network, such ``confusion regions'' can neither symbolize the target nor do not belong to the target. Intuitively, CAM could be enlarged if the pixels of ``confusing regions'' could be reassigned to the right foreground/background areas, as shown by  Figure~\ref{motivation-figure}(b). To drive CAM from coarse-grained to fine-grained, we turn to the complementary learning system, one of the famous frameworks in agent learning theory, to help CAM reassign these confusing pixels.

\paragraph{Complementary Learning System.} 

We consider the perspective of how our brain learns to address WSSS. Figure~\ref{motivation-figure} shows this process according to CLS~\cite{CLS1995, CLS2016} theory. Since bird heads appear in most bird images, our neocortex associates the shape of their heads with birds and pays special attention to heads in images. The association is not entirely exact but effective. After several practical trials, the brain expertly picks out the birds head and excludes the sky. In this process, the hippocampus completes more details of the head, causing attention to spread around. In this way, we learn to distinguish birds from a collection of bird images without pixel-level labels. By analogy with CLS, CAM can be considered to be inadequately trained as in the stage of Figure~\ref{motivation-figure}\ding{172}. Our goal is to build a learning mechanism to adequately train CAM. Therefore, we propose the General-Specific Learning Mechanism (GSLM). We develop the General Learning Module (GLM) to act as the neocortex, which generates the coarse-grained CAM and further processes it to serve as a general localization representation of the target. The Specific Learning Module (SLM) is constructed to complete the pattern complement process similar to the hippocampus, which receives guidance from the representations and reactivates non-discriminating features. In recent years, the inspiration of CLS theory on Deep Neural Networks benefits many fields of Artificial Intelligent (e.g. Deep Q-Network\cite{CLSonDQN}). It makes a lot of sense to introduce CLS into WSSS.

\begin{figure*}[tb]
  \centering
  \includegraphics[width=1.0\textwidth]{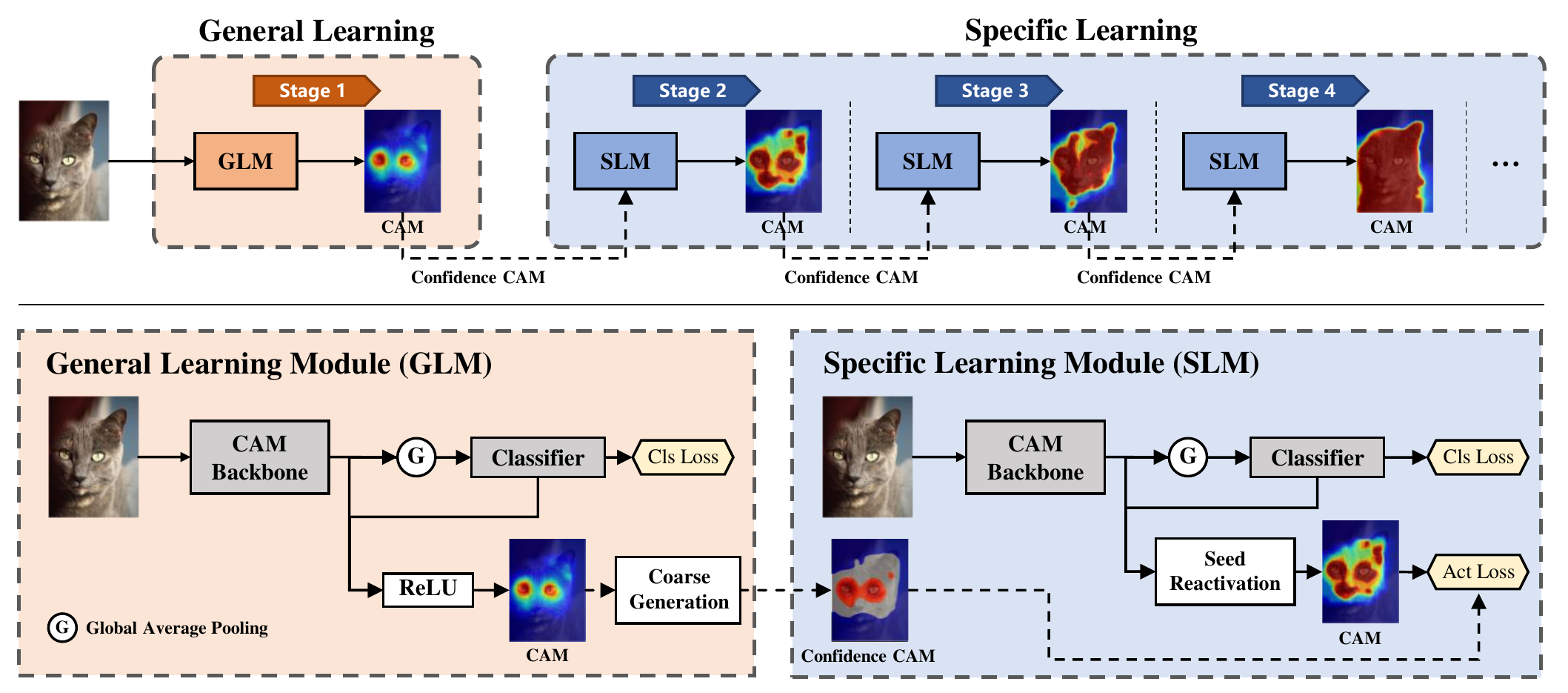}
  \caption{The GSLM framework. Firstly, GSLM trains GLM with image-level labels to generate Confidence CAM. Secondly, SLM is trained with the supervision of image-level labels and Confidence CAM. The CAM is improved after going through GLM and multiple stages of SLM.}
  \label{framwork}
\end{figure*}

\section{Methodology}
\label{methodology}

This section introduces the proposed General-Specific Learning Mechanism (GSLM). In Section~\ref{GSLM}, we illustrate the overall framework of GSLM. Section~\ref{gl} introduces the General Learning Module (GLM), which extracts Confidence CAM with Coarse Generation. Section~\ref{dl} introduces the Specific Learning Module (SLM), into which Seed Reactivation and activation loss are integrated to support pixel-level training and confusion reassignment.

\subsection{Overall framework of GSLM}
\label{GSLM}

GSLM represents a novel training process for a CAM generating network, which consists of a General Learning stage and multiple Specific Learning stages. At each stage, the CAM generating network is wrapped as GLM or SLM, which does not change the topology of the network but affects gradient propagation. Figure~\ref{framwork} shows the GSLM process and the structure of GLM and SLM. Firstly, the network at Stage 1 is wrapped as GLM, which is trained with image-level labels for producing coarse-grained CAM. The coarse-grained will later be refined into Confidence CAM. Secondly, the network at stage 2 is wrapped as SLM, with weights shared from GLM and readjusted under the extra pixel-level supervision of Confidence CAM. The Specific Learning stage is repeated several times, and specifically, the provider of Confidence CAM and shared weights will be the previous SLM instead of GLM.

As shown by Figure~\ref{framwork}, the CAM generation network consists of two branches: classification loss calculation and CAM generation. Structurally, GLM retains the structure of the original CAM generating network and introduces the Coarse Generation for Confidence CAM generation. SLM uses Seed Reactivation to take over the CAM generation branch of the network and supports activation loss calculation.

\subsection{General Learning Module (GLM)}
\label{gl}

As the first stage, GLM aims to generate localization knowledge representation as provision of pixel-level supervision for subsequent stages. Despite carrying localization information, CAM is not suitable for direct use as supervision due to its incomplete object information~\cite{DRS, CPN}. We focus on the trusted local of CAM and propose the Coarse Generation algorithm to generate Confidence CAM, an improved localization map that reinforces salient regions and remarks confusion regions, as supervision. 

\paragraph{Confidence CAM.}

A Confidence CAM indicates three areas with three values, where salient foreground is set to 1, salient background is set to 0, and rest confusion area is set to -1 as the flag to be ignored. The visualization of a sample Confidence CAM is shown as input of SLM in Figure~\ref{framwork}.

\paragraph{Coarse Generation.}

For generating Confidence CAM, we introduce thresholds $\theta_{fg}$ and $\theta_{bg}$ to divide CAM into salient foreground ($\geq \theta_{fg}$), salient background ($<\theta_{bg}$), and confusion area. In addition, conditional random fields (CRF)~\cite{CRF}, a widely used non-learnable CAM refinement method, is applied to further refine the salient regions, introducing the boundary constraint information for such knowledge representations. We define a confidence mapping as $g$ and CRF as $\mathcal{R}: \mathbb{R}^{C \times H \times W} \rightarrow \mathbb{R}^{C \times H \times W}$. For the $c$-th class, Confidence CAM $N_c$ can be generated from given CAM $M_c$ as follows.

\begin{gather}
\label{coarse-generation}
N_{c} = \mathcal{R}(g(M_c)),\ \ \ \  g(x)=\left\{
\begin{array}{ll}
1   &, x \geq \theta_{fg}\\
0   &, x < \theta_{bg}\\
-1  &, \text{else}\\
\end{array} \right..
\end{gather}

\subsection{Specific Learning Module (SLM)}
\label{dl}

SLM aims to readjust the connection weights of CAM generating network with Confidence CAM for Specific Learning. Specifically, Specific Learning refers to consolidating the object patterns of salient regions learned and discovering new patterns to complement the attention of CAM. Therefore, we introduce activation loss, which provides pixel-level supervision with the guiding of Confidence CAM. And we propose Seed Reactivation to help the network activate more regions, free from the crushing effects of salient regions. Besides, the classification loss of GLM is retained for consolidating the learned patterns.

\paragraph{Seed Reactivation}

The conventional CAM is generated by global maximum normalization (Eq. \ref{cam}), in which the maximum pixel value may be located in confusion regions of Confidence CAM and thus be ignored, resulting in gradients that cannot be transmitted back. To address the issue, we introduce a bounded ReLU-k function $\psi^k$ and redefine the generation of CAM (Eq.~\ref{cam}) in SLM as,

\begin{gather}
\label{wam}
M'_{c} = \frac{\psi^k(f(x)_c)}{k},\ \ \ \  \psi^k(x)=\left\{
\begin{array}{ll}
k  &, x \geq k\\
0  &, x < 0\\
x  &, \text{else}\\
\end{array} \right..
\end{gather}

$\psi^k$ provides a limited upper bound for the activation values (logits of the feature maps) and thus fix the problem of narrow activation regions caused by too high activation values in the salient regions in conventional CAM.

\paragraph{Activation Loss}

The activation loss narrows the gap between the improved CAM $M'$ produced by SLM and Confidence CAM $N$ produced by GLM (or SLM in the previous stage). Specifically, we ignore the non-discriminating regions (-1 in Confidence CAM). By measuring the gap with the smooth L1 norm, the activation loss is defined as,

\begin{gather}
\label{act_loss}
L_\text{act}=||M'_i - \bm{N}_i||_1,\ i \in \{x|\bm{N}_x \geq 0, x \in \mathbb{R}^{H \times W}\}.
\end{gather}

The total loss function of SLM is then defined as follows, 

\begin{gather}
\label{total_loss}
L=L_\text{cls} + \alpha L_\text{act},
\end{gather}

where $\alpha$ denotes the balance factor between classification loss and activation loss.

\section{Experiments}
\label{Experiments}

\subsection{Experimental Setting}

\paragraph{Dataset and Evaluation Metrics.}

We performed experiments on the PASCAL VOC 2012 dataset~\cite{voc}, a visual object class challenge with 20 categories built for real scenes. It contains a total of 10,582 training images, of which 1,464 images have pixel-level labels and the rest are bounding box labels, 1,449 validation images with pixel-level labels, and 1,456 test images. We also verify our method using the MS COCO 2014 dataset~\cite{coco} which contains 82,783 train and 40,504 validation images with 81 categories. For each dataset, we train our network with the training images and image-level classification labels only, and evaluate the pseudo-masks with pixel-level labels. The fully supervised semantic segmentation network is trained with the pseudo-masks and evaluated on PASCAL VOC 2012 validation and test set, and MS COCO validation set, respectively. Performance is evaluated by the mean intersection-over-union (mIoU). The evaluation on PASCAL VOC 2012 $test$, without any annotation, is obtained from the official PASCAL evaluation server.

\begin{table}
  \caption{Comparison of pseudo-masks on PASCAL VOC 2012 $train$ set and MS COCO 2014 $train$ set in mIoU (\%). Seed: accuracy of CAM. CRF: accuracy of CAM with CRF. Mask: accuracy of CAM with PSA/IRN. \dag denotes the re-implemented results.}
  \label{seed-contrast}
  \centering
  \setlength{\tabcolsep}{1.35mm}{
  \begin{tabular}{lccccc}
    \toprule
    \multicolumn{1}{c}{\multirow{2}{*}{Method}} &
    \multicolumn{3}{c}{VOC2012} & \multicolumn{2}{c}{COCO} \\ 
    \cmidrule(lr){2-4} \cmidrule(lr){5-6}
    & Seed & CRF  & Mask & Seed & Mask \\
    \midrule
    PSA $_\text{CVPR'18}$\cite{PSA}  & 48.0 & -    & 61.0 & - & - \\
    IRN $_\text{CVPR'19}$\cite{IRN}  & 48.8 & 54.3 & 66.3 & 32.5$^\dag$ & 38.4$^\dag$ \\
    CONTA $_\text{NIPS'20}$\cite{CONTA}  & 48.8 & -    & 67.9 & 28.7 & 35.2 \\
    SEAM $_\text{CVPR'20}$\cite{SEAM}  & 55.4 & 56.8 & 63.6 & 25.1 & 31.5 \\
    CPN $_\text{ICCV'21}$\cite{CPN}    & 57.4 & -    & 67.8 & -    & -    \\
    AdvCAM $_\text{CVPR'21}$\cite{AdvCAM} & 55.6 & 62.1 & 69.9 & -    & -    \\
    AMR $_\text{AAAI'22}$\cite{AMR}    & 56.8 &      & 69.7 & -    & -    \\
    CLIMS $_\text{CVPR'22}$\cite{CLIMS}& 56.6 & -    & 70.5 & -    & -    \\
    SIPE $_\text{CVPR'22}$\cite{SIPE}  & 58.6 & 64.7 & -    & -    & - \\
    W-OoD $_\text{CVPR'22}$\cite{W-OoD}& 59.1 & 65.5 & 72.1 & -    & -    \\
    L2G $_\text{CVPR'22}$\cite{L2G}    & 59.1 & 65.5 & 72.1 & -    & -    \\
    \midrule
    GSLM (Ours)        & 67.5 & 69.2 & 72.4 & \textbf{43.3} & \textbf{45.0} \\
    GSLM$^+$ (Ours)    & \textbf{70.7} & \textbf{73.4} & \textbf{75.1} & 41.4 & 43.7 \\
    \bottomrule
  \end{tabular}
  }
\end{table}

\paragraph{Implementation details.}

We use an ImageNet~\cite{ImageNet} pretrained ResNet50~\cite{resnet} as the CAM backbone in GSLM and the classification layer is a $1 \times 1$ convolution with output channels adapted to the number of categories in the dataset. In Seed Reactivation, we adopt $\psi^k$ for activation values limitation as in Eq.~(\ref{wam}), where $k$ is set to 6. The balance factor $\alpha$ in Eq.~(\ref{total_loss}) is set to 0.5. The thresholds $\theta_{fg}$ and $\theta_{bg}$ in Coarse Generation are set to 0.30 and 0.05, respectively. The number of iterations of SLM is set to 3 for GSLM. We adopt stochastic gradient descent (SGD) for optimization, with different optimization settings for GLM and SLM. The learning rate is initialized to 1.0 for GLM and 0.01 for SLM, decreased at each iteration with polynomial decay~\cite{polynomial}, and further decreased by $1/10$ at the ResNet50 backbone. The batch size is 16, the number of training epochs is 5, and the weight decay is 0.001. The remaining optimization settings are the same as in \cite{IRN}. Moreover, we build a stronger baseline GSLM$^+$ by replacing GLM in GSLM with IRN~\cite{IRN} and keeping SLM the same but with a single iteration. All other settings are the same as GSLM.

\subsection{State-of-the-arts Comparison}
\label{comp}

\begin{figure*}
  \centering
  \includegraphics[width=\textwidth]{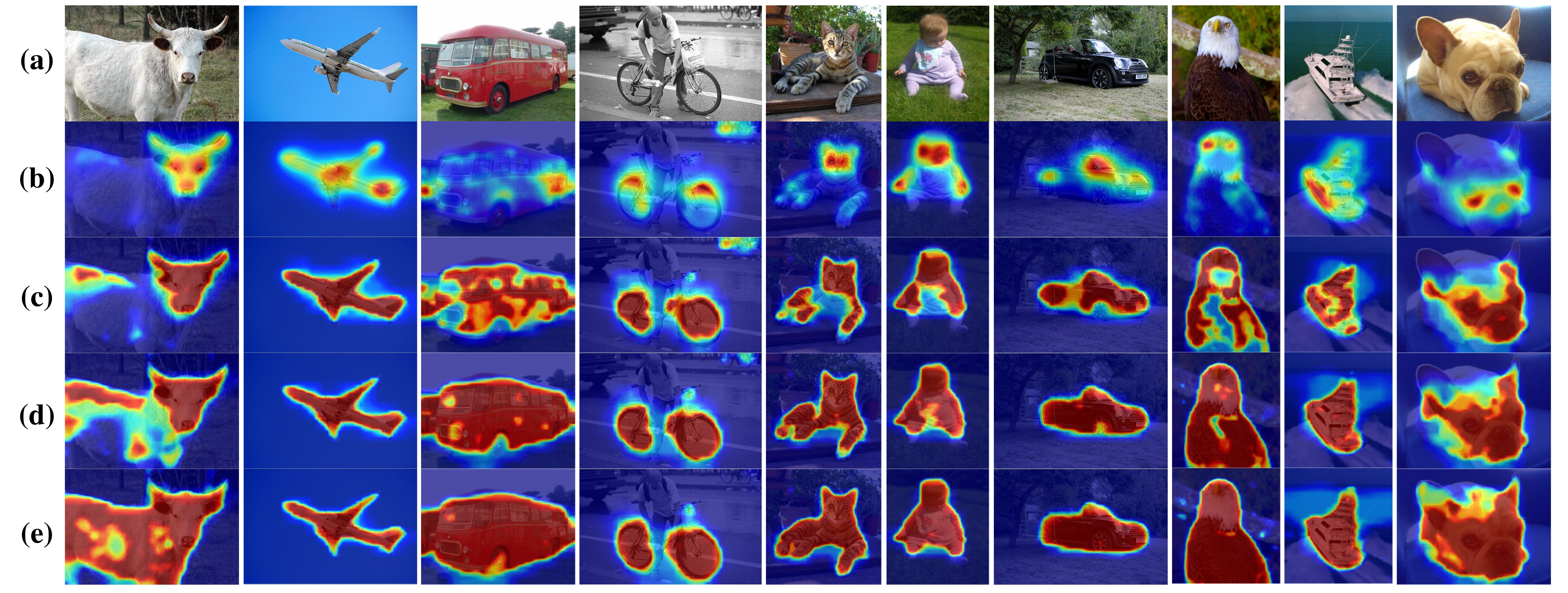}
  \caption{Visualizations of CAM produced by GSLM at different stages on the PASCAL VOC 2012 $train$ set. (a) Original image. (b) CAM in Stage 1 (equivalent to baseline). (c) CAM after Stage 2. (d) CAM after Stage 3. (e) CAM after Stage 4. With the progress of iterations, GSLM activates broader and more accurate regions.}
  \label{show_cam}
\end{figure*}

\paragraph{Improvements on CAM.}

Table~\ref{seed-contrast} reports the performance comparison of CAM seed and pseudo-masks produced by our method and by existing WSSS methods. By convention, pseudo-masks are refined from CAM seeds using PSA~\cite{PSA} or IRN~\cite{IRN}. Without refinement, our GSLM reports an excellent result of 67.5\% mIoU on the PASCAL VOC 2012 $train$ set, outperforming other methods. With the refinement of CRF and IRN (shown as ``Mask" in Table~\ref{seed-contrast}), GSLM also leads the way and generates pseudo-masks with mIoU of 69.2\% mIoU and 72.4\% mIoU, respectively. Figure~\ref{show_cam} shows some samples of CAM produced by GSLM. Moreover, we set up GSLM$^+$ by replacing the initial seed source of GSLM from CAM to IRN (reporting 66.3\% mIoU in Table~\ref{seed-contrast}), achieving a greater performance breakthrough of 70.7\% mIoU (for CAM seed), 73.4\% mIoU (for refinement with CRF) and 75.1\% mIoU (for pseudo-masks), respectively. Compared with the baseline method IRN, GSLM+ provides an improvement of 8.8\% mIoU and is thus the first method to produce CAM seeds with an accuracy exceeding 70\% mIoU. Furthermore, GSLM also performs well on the challenging MS COCO 2014 $train$ set, reporting 43.3\% mIoU for CAM seed and 45.0\% mIoU for pseudo-masks.

\paragraph{Improvements on segmentation results.}

\begin{table}
  \caption{Comparison of state-of-the-art WSSS methods on PASCAL VOC 2012 in mIoU (\%). Marks of supervision (Sup.) denote image-level labels ($\mathcal{I}$), saliency maps ($\mathcal{S}$) and texts ($\mathcal{T}$). All results are based on ResNet backbone.}
  \label{sota}
  \centering
  \setlength{\tabcolsep}{1.30mm}{
  \begin{tabular}{lcccc}
    \toprule
    Method              & Pub.   & Sup.             & Val  & Test \\
    \midrule
    FickleNet~\cite{FickleNet} & CVPR19 & $\mathcal{I+S}$  & 64.9 & 65.3  \\
    ICD~\cite{ICD} & CVPR20  & $\mathcal{I+S}$  & 67.8 & 68.0  \\
    AuxSegNet~\cite{AuxSegNet} & ICCV21   & $\mathcal{I+S}$  & 69.0 & 68.6  \\
    CLIMS~\cite{CLIMS} & CVPR22   & $\mathcal{I+T}$    & 70.4 & 70.0  \\
    NSROM~\cite{NSROM} & CVPR21  & $\mathcal{I+S}$  & 70.4 & 70.2  \\
    L2G~\cite{L2G} & CVPR22  & $\mathcal{I+S}$  & 72.1 & 71.7  \\
    \midrule
    IRN~\cite{IRN} & CVPR19   & $\mathcal{I}$    & 63.5 & 64.8  \\
    SEAM~\cite{SEAM} & CVPR20   & $\mathcal{I}$ & 64.5 & 65.7  \\
    CONTA~\cite{CONTA} & NIPS20  & $\mathcal{I}$  & 66.1 & 66.7  \\
    AdvCAM~\cite{AdvCAM} & CVPR21   & $\mathcal{I}$    & 68.1 & 68.0  \\
    CPN~\cite{CPN} & ICCV21   & $\mathcal{I}$    & 67.8 & 68.5  \\
    SIPE~\cite{SIPE} & CVPR22   & $\mathcal{I}$    & 68.8 & 69.7  \\
    W-OoD~\cite{W-OoD} & CVPR22   & $\mathcal{I}$    & 69.8 & 69.9  \\

    Kho \emph{et, al.}~\cite{kho} & PR22   & $\mathcal{I}$    & 69.5 & 70.5  \\
    \midrule
    GSLM (Ours)      &    & $\mathcal{I}$    & 69.4 & 69.7  \\
    GSLM$^+$ (Ours) &    & $\mathcal{I}$    & \textbf{70.6} & \textbf{70.6}  \\
    \bottomrule
  \end{tabular}
  }
\end{table}

\begin{table}
  \caption{Comparison of state-of-the-art WSSS methods on MS COCO 2014 in mIoU (\%). Marks of supervision (Sup.) denote image-level labels ($\mathcal{I}$) and saliency maps ($\mathcal{S}$). Methods marked by * use VGG backbone, and the other marked by \dag use ResNet backbone.}
  \label{sota-coco}
  \centering
  \setlength{\tabcolsep}{1.30mm}{
  \begin{tabular}{lccc}
    \toprule
    Method              & Pub.   & Sup.             & Val \\
    \midrule
    $^*$ADL~\cite{ADL} & PAMI20 & $\mathcal{I+S}$  & 30.8  \\
    $^\dag$AuxSegNex~\cite{EPS} & ICCV21 & $\mathcal{I+S}$  & 33.9  \\
    $^\dag$EPS~\cite{EPS} & CVPR21 & $\mathcal{I+S}$  & 35.7  \\
    $^*$L2G~\cite{L2G} & CVPR22 & $\mathcal{I+S}$  & 42.7  \\
    \midrule
    $^*$SEC~\cite{SEC} & ECCV16 & $\mathcal{I}$  & 22.4  \\
    $^\dag$IRN~\cite{IRN} & CVPR19 & $\mathcal{I}$  & 32.6  \\
    $^\dag$CSE~\cite{CSE} & ICCV21 & $\mathcal{I}$  & 36.4  \\
    $^*$RCA~\cite{RCA} & CVPR22 & $\mathcal{I}$  & 36.8  \\
    \midrule
    $^\dag$GSLM (Ours)      &   & $\mathcal{I}$    & \textbf{40.9}  \\
    \bottomrule
  \end{tabular}}
\end{table}

\begin{figure}
  \centering
  \includegraphics[width=0.47\textwidth]{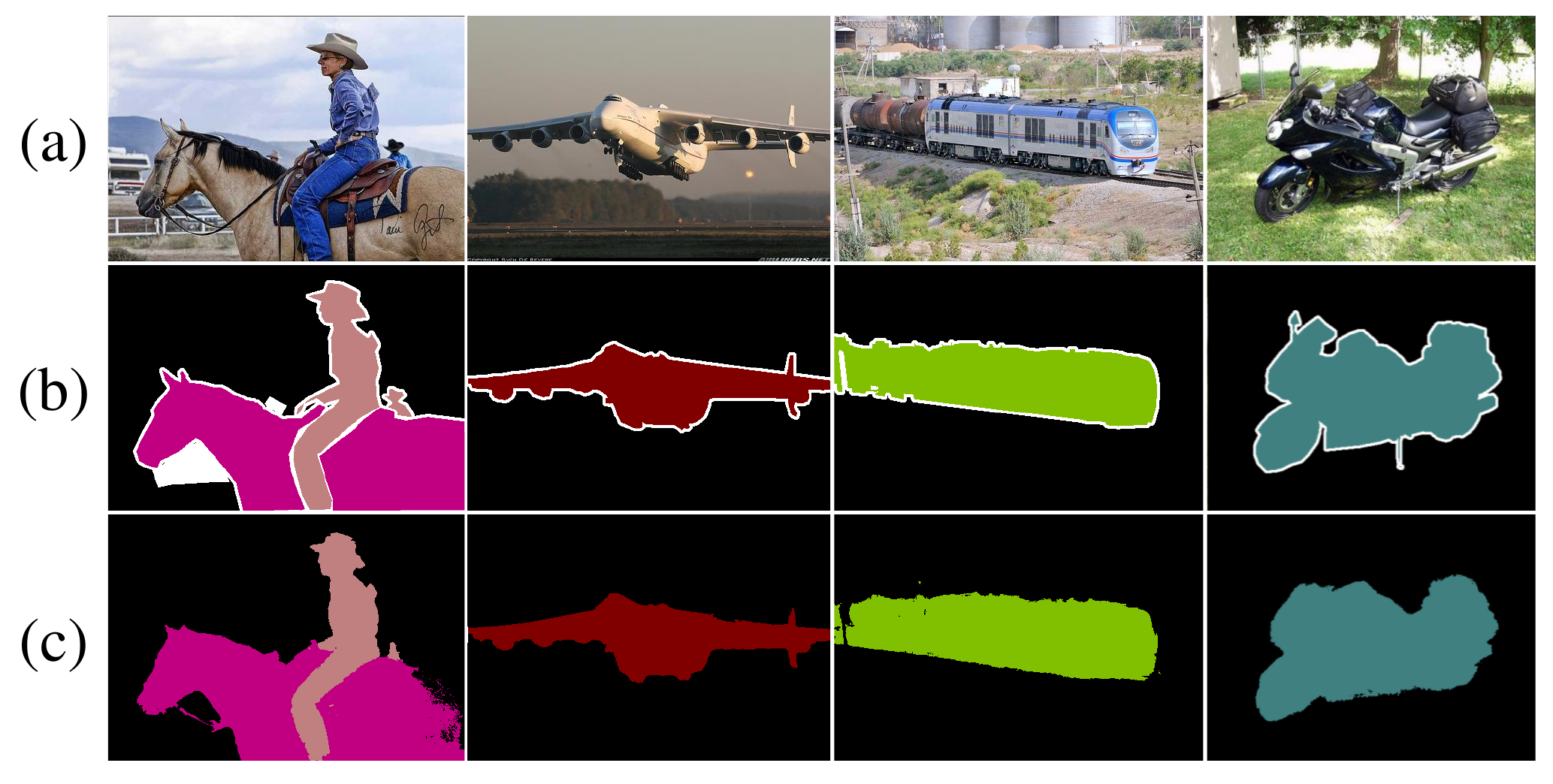}
  \caption{Qualitative results on PASCAL VOC 2012 $val$ set. (a) Input Images. (b) Ground-truths. (c) results of our GSLM$^+$ (w/ CRF).}
  \label{show_seg}
\end{figure}

To apply our GSLM in semantic segmentation, we train DeepLabV2-ResNet~\cite{DeeplabV2} with the pseudo-masks generated by GSLM and GSLM$^+$, respectively. Table~\ref{sota} shows the comparison of the performance of our method and existing state-of-the-art WSSS methods on the PASCAL VOC 2012 $val$ and $test$ benchmarks. Our GSLM achieves 69.4\% mIoU on $val$ and 69.7\% mIoU on $test$. With the help of the seeds provided by IRN, GSLM$^+$ achieves 70.6\% on $val$ and 70.6\% on $test$, an improvement of 7.1\% and 5.8\% over IRN. Our GSLM$^+$ outperforms all existing methods with only supervision of image-level labels and most methods with extra supervision of saliency maps, representing new state-of-the-art performance on the PASCAL VOC 2012 benchmark. Beside, Table~\ref{sota-coco} shows that our GSLM is also efficient for challenging COCO data sets and achieve 40.9\% mIoU, which is ahead of existing methods. Figure~\ref{show_seg} shows some qualitative samples of semantic segmentation results of our GSLM$^+$, validating its effectiveness.

\subsection{Ablation Study}

\paragraph{Effectiveness on different baselines.}

In GSLM, GLM generates CAM seeds for Coarse Generation by applying a conventional ResNet50 network. GSLM increases the baseline CAM seeds from 48.6\% mIoU to 67.5\% mIoU, showing excellent performance. GSLM could also be applied to various WSSS frameworks. As shown in Table~\ref{refine-effectiveness}, GSLM greatly improves the accuracy of the CAM seeds for various WSSS methods by adopting the premium initial seeds produced by them. Specifically, it boosts SEAM to 67.6\% mIoU (+12.2\% mIoU), CPN to 66.9\% (+9.6\% mIoU), and AdvCAM to 68.0\% (+12.5\% mIoU), all significantly ahead of existing state-of-the-art methods, as shown in Table~\ref{seed-contrast}. In particular, GSLM improves the performance of IRN up to 70\% mIoU, representing a new milestone in the accuracy of CAM in WSSS.

\begin{table}
  \caption{The refinement effectiveness of our GSLM with initial seed produced by different baseline in mIoU (\%) on PASCAL VOC 2012.}

  \label{refine-effectiveness}
  \centering
  \begin{tabular}{lcc}
    \toprule
    Method & Seed & +GSLM\\
    \midrule
    CAM\cite{CAM}   & 48.6 & 67.5\textcolor{red}{$_{+18.9}$} \\
    SEAM\cite{SEAM} & 55.4 & 67.6\textcolor{red}{$_{+12.2}$} \\
    CPN\cite{CPN}   & 57.3 & 66.9\textcolor{red}{$_{+\ 9.6\ }$} \\
    AdvCAM\cite{AdvCAM} & 55.5 & 68.0\textcolor{red}{$_{+12.5}$} \\
    IRN\cite{IRN}   & 66.5 & 70.7\textcolor{red}{$_{+\ 4.2\ }$} \\
    \bottomrule
  \end{tabular}

\end{table}

\paragraph{Effect of main modules.}

To evaluate the effect of each part of GSLM, we remove one of Coarse Generation (CR), Seed Reactivation (SR), and classification loss ($L_{cls}$) in Eq~(\ref{cls_loss}), respectively, and calculate the mIoU of GSLM, as shown in Table~\ref{ablation}. When CR or SR is removed, GSLM loses its usefulness and the performance drops to 49.3\% mIoU and 49.5\% mIoU. When $L_{cls}$ is removed, performance decreased by 1.5\% compared to the complete GSLM. In other words, CR and SR are critical modules, and $L_{cls}$ improves the performance of GSLM. Moreover, by applying SLM iteratively, the result further achieves a 7.4\% improvement.

\paragraph{Effect of boundary constraint in coarse generation.}

\begin{figure}
  \centering
  \includegraphics[width=0.45\textwidth]{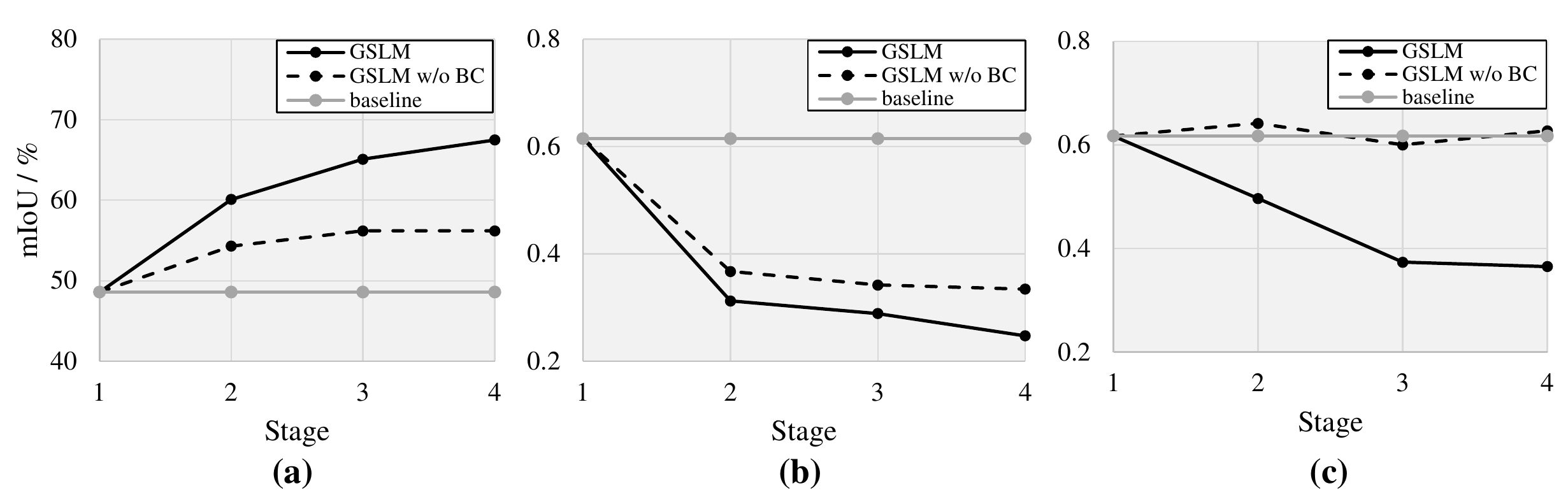}
  \caption{Comparison of GSLM with and without boundary constraint (BC) on (a) accuracy, (b) under-activation, and (c) over-activation in different stages. For (b), the lower the degree of under-activation, the more regions of objects are activated by the network. For (c), the lower the degree of over-activation, the more accurate the network activation area is.}
  \label{rc-ablation}
\end{figure}

\begin{figure}
  \centering
  \includegraphics[width=0.45\textwidth]{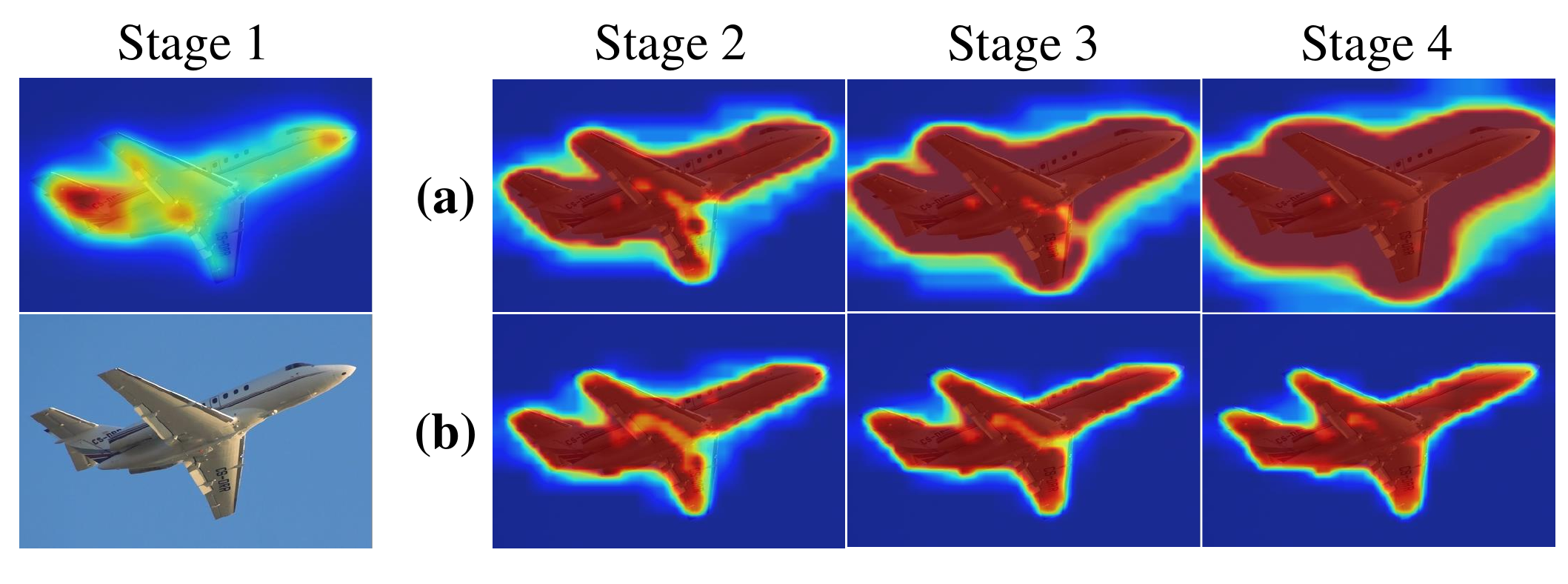}
  \caption{Comparison of CAM changes at different stages of GSLM without (a) and with (b) boundary constraint (BC).}
  \label{rc-ablation-pic}
\end{figure}

As described in Section~\ref{gl}, CRF is used in Coarse Generation for preliminary refinement of the Confidence CAM. Furthermore, we believe that CRF can introduce boundary information to CAM for advantage of reactivation in SLM. To validate this view, we compare GSLM with and without boundary constraint (by adopting or not adopting CRF in Coarse Generation) in terms of accuracy (mIoU), under-activation, and over-activation, respectively. The results are shown in Figure~\ref{rc-ablation}. We evaluate the degree of under-activation and over-activation by following metrics proposed in~\cite{SEAM}:

\begin{gather}
m_{under} = \frac{1}{C-1}\sum^{C-1}_{c=1}\frac{FN_c}{TP_c}, \\
m_{over} = \frac{1}{C-1}\sum^{C-1}_{c=1}\frac{FP_c}{TP_c},
\end{gather}
where $C$ denotes the number of categories with background, $TP_c$ denotes the pixel number of the true positive prediction of class $c$, $FP_c$ and $FN_c$ denote false positive and false negative, respectively.

In Figure~\ref{rc-ablation}(a), GSLM with BC demonstrates better performance than GSLM without BC in all rounds. To find out the reason for this phenomenon, Figure~\ref{rc-ablation}(b) and Figure~\ref{rc-ablation}(c) calculate the changes of the degree of under-activation and over-activation of the two settings of GSLM in the iteration process, respectively. Specifically, both of them show similar performance in the optimization of under-activation. However, in the absence of boundary constraints, GSLM fails to control the over-activation degree, as shown in Figure~\ref{rc-ablation}(c), the over-activation degree of GSLM without BC is close to the baseline. Figure~\ref{rc-ablation-pic} shows the visualization of this phenomenon. The activated region of GSLM without BC to the airplane in the image is excessively diffused to the background region with the progress of iteration, which hinders the improvement of the accuracy. In conclusion, the under-activation problem of CAM is reduced due to the reactivation of non-discriminating regions by the SLM of GSLM. Moreover, the over-activation problem is reduced by introducing boundary constraints in Coarse Generation.

\begin{table}
  \caption{The ablation study for each part of GSLM on PASCAL VOC 2012. CR: Coarse Generation. SR: Seed Reactivation. $L_{cls}$: classification loss in Eq~(\ref{cls_loss}). Iter: Iteration of SLM.}
  \label{ablation}
  \centering
  \begin{tabular}{cccccc}
    \toprule
    baseline 
    & CR
    & SR
    & $L_{cls}$
    & Iter & mIoU (\%)\\
    \midrule
    $\checkmark$&            &            &            &            & 48.6 \\
    \midrule
    $\checkmark$&            &$\checkmark$&$\checkmark$&            & 49.3 \\
    $\checkmark$&$\checkmark$&            &$\checkmark$&            & 49.5 \\
    $\checkmark$&$\checkmark$&$\checkmark$&            &            & 58.6 \\
    $\checkmark$&$\checkmark$&$\checkmark$&$\checkmark$&            & 60.1 \\
    \midrule
    $\checkmark$&$\checkmark$&$\checkmark$&$\checkmark$&$\checkmark$& 67.5 \\
    \bottomrule
  \end{tabular}
\end{table}

\paragraph{Effect of $\bm{\theta_{bg}, \theta_{fg}, k}$, and number of iterations.}

\begin{figure}[tb]
  \centering
  \includegraphics[width=0.9\textwidth]{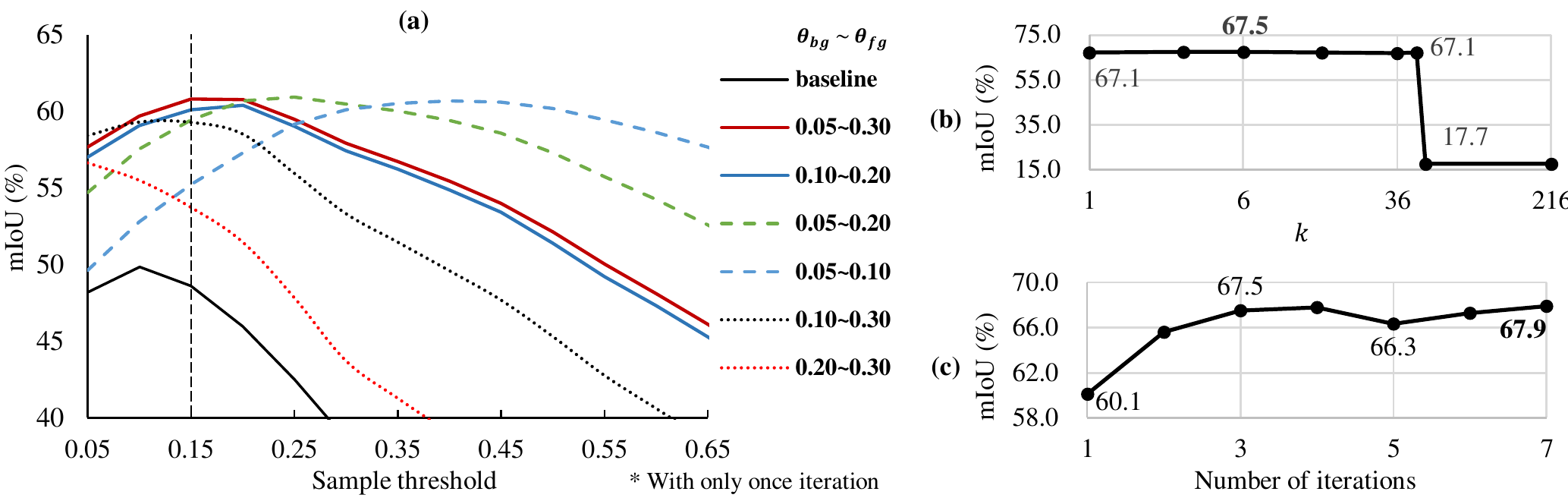}
  \caption{The CAM performance of GSLM with different (a) $\theta_{bg}, \theta_{fg}$, (b) $k$, and (c) number of iterations on PASCAL VOC 2012.}
  \label{parameter}
\end{figure}

Figure~\ref{parameter} reports the effects of hyper-parameters on the CAM accuracy of GSLM, respectively.

\begin{itemize}[leftmargin=*]

\item $\theta_{bg}, \theta_{fg}$: Figure~\ref{parameter}(a) shows the CAM performance curve of GSLM for difference $\theta_{bg}, \theta_{fg}$. The sampling threshold is 0.15, with which baseline achieves 48.6\% mIoU. When $(\theta_{bg}, \theta_{fg})$ is $(0.05,0.3)$, the best threshold of GSLM coincides with the sampling threshold. However, decreasing $\theta_{fg}$ (dashed line item) or increasing $\theta_{bg}$ (dotted line item) makes the best threshold deviate from the sample threshold, which is not conducive to more iteration of GSLM. Narrowing the interval of $\theta_{bg}, \theta_{fg}$ (blue item) will slightly reduce performance.

\item $k$: Figure~\ref{parameter}(b) shows that GSLM is insensitive to $k$ when $k \leq 45$, and GSLM diverges when $k > 45$. This is due to the excessively large value of $k$, resulting in an excessively high output of the loss function Eq.(\ref{total_loss}). 

\item number of iterations: Figure~\ref{parameter}(c) shows that the accuracy of GSLM is stable after 3 iterations.

\end{itemize}

\section{Conclusion}

In this paper, we aim to address WSSS by exploring and exploiting the Complementary Learning System (CLS) for CAM, which is mostly confined to small discriminating object areas. Specifically, we proposed General-Specific Learning Mechanism (GSLM) to help drive CAM advance in a fine-grained way. GSLM consists of General Learning Module (GLM) and Specific Learning Module (SLM). Specifically, GLM extracts localization information with boundary constraints from images and stores it as Confidence CAM. SLM organizes region reassigning guided by Confidence CAM, which reinforces salient regions and remarks confusion regions. Experimental results validate that our method achieves new state-of-art performance.

\bibliographystyle{unsrt}  
\bibliography{references}

\end{document}